 \documentclass[pmlr,twocolumn,10pt]{jmlr} 

\usepackage{bbm}
\usepackage{booktabs}
\usepackage[load-configurations=version-1]{siunitx} 

\theorembodyfont{\upshape}
\theoremheaderfont{\scshape}
\theorempostheader{:}
\theoremsep{\newline}

\jmlryear{2021}
\jmlrworkshop{Machine Learning for Health (ML4H) 2021} 

 \title[HAD-Net: Hybrid Attention-based Diffusion Network for Glucose Level Forecast]{HAD-Net: Hybrid Attention-based Diffusion Network \titlebreak for Glucose Level Forecast}

\author{%
\Name{Quentin Blampey} \Email{quentin.blampey@gmail.com}\\
\Name{Mehdi Rahim} \Email{mehdi.rahim@airliquide.com}\\
\addr Computational \& Data Science, Air Liquide R\&D. Innovation Campus Paris, 78350, Les Loges-en-Josas, France
}

\begin{document}

\maketitle

\begin{abstract}
Data-driven models for glucose level forecast often do not provide meaningful insights despite accurate predictions. Yet, context understanding in medicine is crucial, in particular for diabetes management. 
In this paper, we introduce HAD-Net: a hybrid model that distills knowledge into a deep neural network from physiological models. It models glucose, insulin and carbohydrates diffusion through a biologically inspired deep learning architecture tailored with a recurrent attention network constrained by ODE expert models.
We apply HAD-Net for glucose level forecast of patients with type-2 diabetes. It achieves competitive performances while providing plausible measurements of insulin and carbohydrates diffusion over time.\end{abstract}

\section{Introduction}
\label{sec:intro}

Diabetes is a disease where glucose level fluctuations are extreme. Insulin inefficiency causes high glucose levels (hyperglycemia) or low glucose level (hypoglycemia). Correct insulin intake and a good knowledge of patient metabolism are key to improving the control of glucose level.
In this context, glucose-level forecast models aim at anticipating hyper/hypoglycemia episodes, and thus, improving diabetes management and patient daily life. Such models predict the glucose-level within the near future (minutes to hours) from glucose level variation, insulin delivery and meal intake.
There is a large variety of models for glucose-level forecast \citep{Woldaregay2019}. They can be split into three categories: \textbf{Physiological models} \citep{ode_bench} are based on expert knowledge to describe patient metabolism. There are many physiological models, from minimal models \citep{Bergman2005} to complex ones \citep{uva_padova}. These models lack flexibility and do not fit well with every patient metabolism, resulting in inaccurate forecasts. Conversely, \textbf{data-driven models} \citep{bglp_gans, oreshkin2020nbeats} usually achieve better results despite a lack of interpretability. \textbf{Hybrid models} combine both approaches and have promising properties of physiological meaningfulness and forecast accuracy. E.g. \cite{miller2020learning} introduces a hybrid approach that injects a deep learning model to augment the UVA-PADOVA model \citep{uva_padova}. They provide a flexible simulation of insulin-glucose dynamics but limited short-term forecasts.
In this work, we present a hybrid approach that injects a physiological model into a neural network for glucose forecast. 
It reproduces the natural diffusion of physiological quantities: it tracks the food in the stomach, going into the gut, and then transforming into blood glucose. Its core mechanism is similar to solving model ODEs, the main difference is that diffusion magnitudes are learned by an attention mechanism. The latter weights are learned via back-propagation. 
The physiological model enforces the model to be physiologically consistent, while the neural network core extracts complex dependencies and features.

\vspace{-0.7mm}
\section{Methods}
\label{sec:methods}

\paragraph{Notations}
\label{sec:notations}

Let $\vec{X}_{1:w} = (\vec{x_1}, \vec{x_2}, .., \vec{x_w}) \in \mathbb{R}^{m \times w}$ be a multidimensional time series of $m$ measured variables. That is, $w$ measures were made at a constant time step $\Delta t$ (5 minutes for our use case). Here, $\vec{X}_{a:b}$ denotes the columns $a$ to $b$ (included) of $\vec{X}$. We consider that $(X_{1, i})_{1 \leq i \leq w}$ represents glucose measures. The other time series can represent any measured variable, e.g. insulin or carbs intake. We aim at predicting the  glucose level for the next $h$ steps $(X_{1, w+i})_{1 \leq i \leq h}$, where $h$ is a provided time horizon. Also, physiological models for diabetes are systems of equations defined by experts that describe the metabolism dynamics related to glucose. Let $\vec{v} = (p_1, p_2, ..., p_K) \in \mathcal{F}(\mathbb{R}_+, \mathbb{R}_+^K)$ be a vector of $K$ time-dependent physiological variables (PVs). The dynamics are defined by a function $g$ and an initial state $\vec{v_0}$ such as $\vec{v}(0) = \vec{v_0}, \quad \frac{d\vec{v}}{dt}(t) = g(\vec{v}(t), t, \vec{\gamma})$ where $\vec{\gamma}$ are patient specific parameters.

\subsection{Contribution}
\label{sec:contribution}

We propose a neural diffusion model for glucose level forecast that uses physiologically inspired constraints.
Our approach represents a physiological model of metabolism as a graph diffusion model.
Flow magnitudes of this diffusion model are learned by an adapted attention mechanism. 
Flow directions are constrained by an adjacency matrix of glucose/insulin/carbs interactions.
We show that such a hybrid approach brings meaningful insights on glucose-insulin interactions and yields competitive predictions of glucose level on real data.

\subsection{Graph Diffusion Physiological Model}
\label{sec:gdpm}

A graph diffusion physiological model (GDPM) is a rewritting of a physiological model (see \appendixref{apd:gdpm}). Each PV corresponds to a graph node, and flows are exchanged along edges. Under some conditions, dynamics can be expressed as the following set of equations, $\forall i \in [1, K], \forall t \geq 0$:
\begin{equation}
\frac{dp_i}{dt}(t) = \sum_{j \mid w_{ji} \neq 0} w_{ji}I_{ji}(t)p_j(t) -\!\!\! \sum_{j \mid w_{ij} \neq 0} I_{ij}(t)p_i(t),
\end{equation}
where $I_{ij}$ represents flow magnitude from node $i$ to $j$, and the edges weights $w_{ij} \in \{1, 0, -1\}$ represents flow direction and its characteristics (constructive or destructive process). The first sum corresponds to incoming flow, while the other sum corresponds to outgoing flow. Now, let $\vec{A} = (w_{ij})_{i, j}$, and $\vec{F} = (I_{ij})_{i, j}$. The previous system of equations can be rewritten as
\begin{equation}\label{eq:gdpm_dynamics}
\frac{d\vec{v}}{dt} = (\vec{A} \odot \vec{F} - Diag((|\vec{A}|^T \odot \vec{F})_{\rightarrow}))\vec{v},
\vspace{-2mm}
\end{equation}
where $\odot$ is the element-wise product, $|\vec{.}|$ the element-wise absolute value, $Diag$ builds a diagonal matrix out of a vector, and $\rightarrow$ builds a vector out of the sum along the rows of a matrix.

\subsection{Model Architecture}
\label{sec:model}

Our model architecture depends on a physiological model. Let's rewrite the latter as a GDPM, from which we extract $\vec{A}$. We consider discrete time steps and denote $\vec{V} = (\vec{v}_t)_{t \geq 1}$ the PVs values at time-steps $t = 1, 2, ..., w+h$. We learn the magnitude matrix $\vec{F}$ with a neural network such that we can compute the PVs increments $\vec{v}_{t+1} - \vec{v}_t$. This step is called diffusion step, and it has similarities with graph diffusion networks \citep{chamberlain2021grand}. Multiple diffusion steps are necessary to infer $\vec{v_w}$ based on $\vec{X}_{1:w}$. Then, the forecast is obtained by composing multiple diffusion steps until we reach the desired horizon $h$.

\begin{figure*}[hbt!]
\centering
\includegraphics[width=\linewidth]{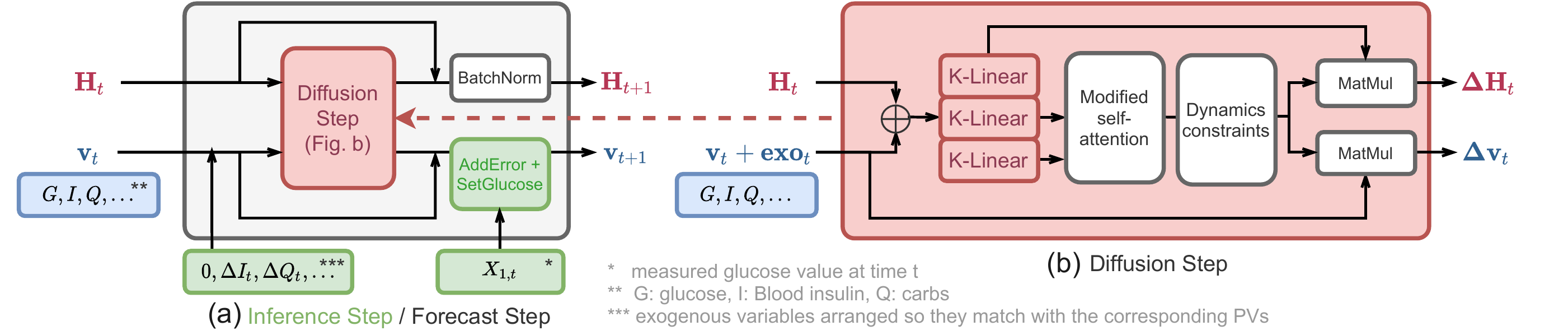}
\caption{\label{fig:core_step}\textbf{Representation of the inference step / forecast step (a).} The forecast step, similar to the inference step, is obtained by removing the green box operations that concern unknown variables during forecast. Both steps update the PVs $\vec{v}$ and the hidden states $\vec{H}$ from time $t$ to $t+1$. The diffusion step (b) is based on an adaptation of self-attention with constraints of an expert model.}
\end{figure*}
\subsubsection{Diffusion step}
\label{sec:diffusion_step}

The core of the model relies on learning the increment to update the PVs from time $t$ to $t+1$, where $1 \leq t \leq w+h-1$. We learn $\vec{F}$ via an attention process and the Euler method applied on \equationref{eq:gdpm_dynamics}:
\begin{equation}
\label{eq:increment}
   \vec{v}_{t+1} - \vec{v}_{t} = (\vec{A} \odot \vec{F} - Diag((|\vec{A}|^T \odot \vec{F})_{\rightarrow}))\vec{v}_{t}\Delta t 
\end{equation}
\paragraph{Adding hidden states}
Every PV corresponds to one single scalar. Computing attention between scalar features would produce too simple dynamics, so we need extra features to capture complex relations and memorize patterns. We consider $K$ hidden features vectors of size $d-1$, where $d$ is called the model size. We associate $\vec{v}_{t}$ with the hidden state matrix $\vec{H}_{t} \in \mathbb{R}^{K \times (d-1)}$. Their concatenation gives an embedding matrix $\vec{E}_t = \vec{v}_t^T || \vec{H}_t$ of size $K \times d$.
\paragraph{Parameter-wise projection into queries, keys, values}
We define $\forall 1 \leq k \leq K, \vec{L_k^Q} $ being affine transformations of $\mathbb{R}^d$. Let's $\vec{Q} := (\vec{e_k}\vec{L_k^Q})_{1 \leq k \leq K}$, whose columns are made from the parameter-wise transformations of the rows of $\vec{E}_t$. Because PVs do not have the same role, this parameter-wise transformation (called $\mathbf{K-Linear}$) is needed. Thus, we have $K$ more matrix to create $\vec{Q}$ than in \citet{vaswani2017attention}. We create $\vec{K}, \vec{V}$ by the same process.

\paragraph{Adapted attention mechanism}
Let $a_0$ be default attention such that $\sigma(a_0)$ is a characteristic time of the dynamics, where $\sigma$ is the sigmoid function. One hour (12 time-steps of $\Delta t = 5$ minutes) is a typical value of insulin effect duration. We can thus take $a_0$ s.t.  $\sigma(a_0) = \frac{1}{12}$. It indicates to the model how fast the system should evolve, e.g. it shouldn't have significant changes every minute. We replaced the softmax with a sigmoid because cross magnitudes should only depend on a pair of variables. It leads to the attention $\vec{F} = \sigma ( a_0 + \frac{\vec{Q}\vec{K^T}}{d})$. Then, we update $\vec{v}_t$ based on \equationref{eq:increment}. We do not use batch normalization on PVs because scaling would introduce physiologically inconsistent values (see \algorithmref{alg:diffusion_step}).

\subsubsection{Inference and forecast algorithms}
\label{sec:inference_step}

Prediction (detailed on \algorithmref{alg:full}) can be decomposed into inference and forecast. They are respectively compositions of inference steps and forecast steps. The forecast step (see \figureref{fig:core_step}) is the diffusion step with extra operations listed below:
\paragraph{AddExogenous} At each time step, we add the exogenous variables measured in $\vec{x}_t$ to the corresponding PVs. For instance, if $X_{2, t}$ corresponds to insulin intake and $V_{2, t}$ corresponds to the blood insulin compartment, then $V_{2, t} \gets V_{2, t-1} + X_{2, t}$.
\paragraph{SetGlucose} As the glucose is known at each time step during inference, we can correct the glucose variable so it doesn't accumulate errors during inference, i.e. $V_{1, t} \gets X_{1, t}$ where $V_{1, t}$ is the glucose PV.
\paragraph{AddError} The model needs to know the glucose level error during inference to adjust itself. We create two additional PVs corresponding to cumulative past errors, denoted $\epsilon^+, \epsilon^-$. They accumulates positive errors depending on the sign of the error $\epsilon_t := X_{1, t} - V_{1, t}$, where $V_{1, t}$ is the glucose PV. That is, $\epsilon_t^+ \gets \epsilon_{t-1}^+ + \epsilon_t \mathbbm{1}_{\mathbb{R}^+}(\epsilon_t)$ and $\epsilon_t^- \gets \epsilon_{t-1}^- -  \epsilon_t\mathbbm{1}_{\mathbb{R}^-}(\epsilon_t)$. We update the number of PVs $K \gets K + 2$ and the matrix $\vec{A}$ meaningfully (details on \appendixref{apd:implementation}).

\subsection{Loss and Back-Propagation}
\label{sec:loss}

We learn our model parameters via back-propagation of a loss $
\mathcal{L} = \mathcal{L}_{MSE} + \alpha_1\mathcal{L}_{\epsilon} + \alpha_2 \mathcal{L}_{nr}
$ where $\alpha_1, \alpha_2$ are weights that are set empirically. $\mathcal{L}_{MSE}$ is the MSE loss applied on the continuous predictions from horizon 1 to $h$. $\mathcal{L}_{\epsilon} := \|\vec{\epsilon^+}\|_2^2 + \|\vec{\epsilon^-}\|_2^2$ penalizes the errors made by the model during inference. Finally, $\mathcal{L}_{nr}$ penalizes not realistic values of $\vec{F}$. Indeed, the values should be consistent with physiological order of magnitude (e.g. insulin sensitivity factor $\in [10, 50]$). This term was defined manually to make sure our model stays consistent with the reality.
\vspace{-4mm}
\section{Experiments and Results}
\label{sec:results}

We show the performance of HAD-Net on real data of diabetic patients. We benchmark HAD-Net against state-of-the-art models, and we analyze it depending on physiological contexts.
We train HAD-Net on a dataset of 17 patients with type-2 diabetes. Each patient has 2 weeks of data records where glucose level is measured every 5 minutes with a continuous glucose monitoring device (CGM). Insulin delivery and meal intakes are also reported. The total dataset contains 59k CGM data points.
We apply a 80\%-20\% train-test split on each time series with 10 repetitions.
\begin{table*}[hbtp!]
  \centering
  \footnotesize
  \caption{\label{tab:metrics}Comparison of predictive models for  on validation set over 10 repetitions (mean$\pm$std).}
    {
  \begin{tabular}{l|cc|cc}
  \toprule
  { \bf Model} & 
  { \bf RMSE (30min)} &
  { \bf RMSE (60min)} &
  { \bf MARD (30min)} &
  { \bf MARD (60min)}\\
  \midrule
  Baseline & 24.3 $\pm$ 0.8 & 38.1 $\pm$ 1.3 & 13.7\% $\pm$ 0.6 & 20.1\% $\pm$ 1.1\\
    Physiological &   25.7 $\pm$ 1.1 & 38.1 $\pm$ 1.4 &   12.4\% $\pm$ 0.8 & 20.5\% $\pm$ 1.4\\
  ARIMA & 23.2 $\pm$ 1.1 & 40.0 $\pm$ 2.0 & 10.7\% $\pm$ 0.6 & 19.3\% $\pm$ 1.2\\
  Ridge & 21.4 $\pm$ 0.7 & 33.4 $\pm$ 1.2 & 12.3\% $\pm$ 0.5 & 18.6\% $\pm$ 1.0 \\
  Gaussian Process &   23.3 $\pm$ 1.5 & 34.5 $\pm$ 1.2 &   12.5\% $\pm$ 2.4 & 19.5 $\pm$ 2.8\\
  CNN-MLP & 17.9 $\pm$ 0.8 & 30.2 $\pm$ 1.3 & 8.7\% $\pm$ 0.6 & 14.6\% $\pm$ 1.0\\
    LSTM &   17.1 $\pm$ 0.6 & \textbf{28.4 $\pm$ 1.1} &   \textbf{8.4\% $\pm$ 0.6} & \textbf{13.8\% $\pm$ 1.1}\\
  GRU & 17.0 $\pm$ 0.6 & 28.4 $\pm$ 1.2 & 8.4\% $\pm$ 0.7 & \textbf{13.8\% $\pm$ 1.1}\\
  HAD-Net \scriptsize (ours) & \textbf{16.9 $\pm$ 1.2} & 28.4 $\pm$ 1.7 & \textbf{8.4\% $\pm$ 0.6} & 13.8\% $\pm$ 1.3\\
  \bottomrule
  \end{tabular}}
\end{table*}
\paragraph{Glucose-level prediction benchmark.}
We compare our model to the persistence model,
linear models (ARIMA, Ridge regression), a physiological model,
and neural networks (CNN-MLP, LSTM, GRU). These models (\appendixref{apd:models}) are trained on all patients, except CNN-MLP that is patient-specific. 
\tableref{tab:metrics} summarizes averaged RMSE (root mean squared error) and MARD (mean absolute relative deviation) on held-out test sets. We report errors for 30 minutes and 60 minutes prediction horizon.
Results show that neural network models outperform the baseline and linear models.
More interestingly, HAD-Net has the lowest RMSE compared to other neural network models.

\begin{figure}[hbt!]
\centering
\includegraphics[width=0.5\textwidth]{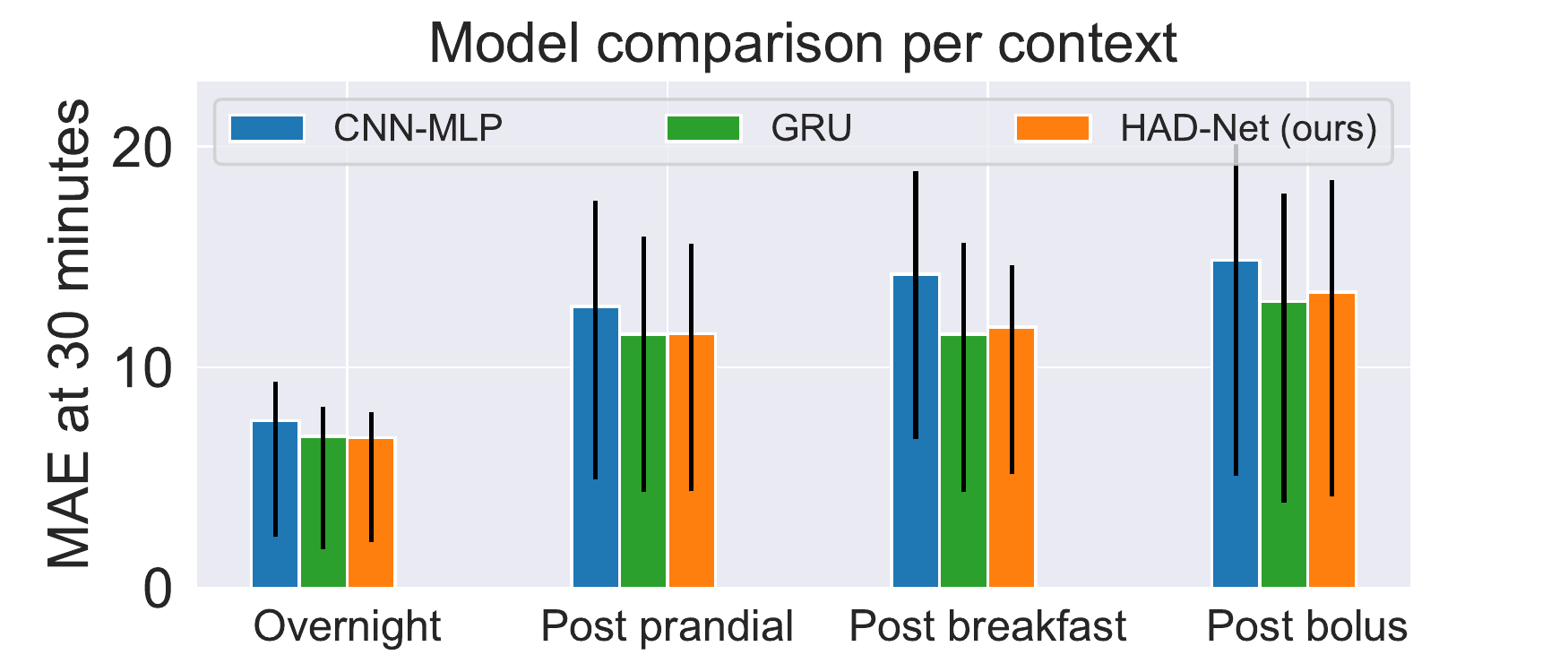}
\caption{\label{fig:context_results}\textbf{Performance comparison of neural networks per context.} Glucose level predictions are computed at 30 minutes.}
\end{figure}

\begin{figure}[hbt!]
\centering
\includegraphics[width=0.5\textwidth]{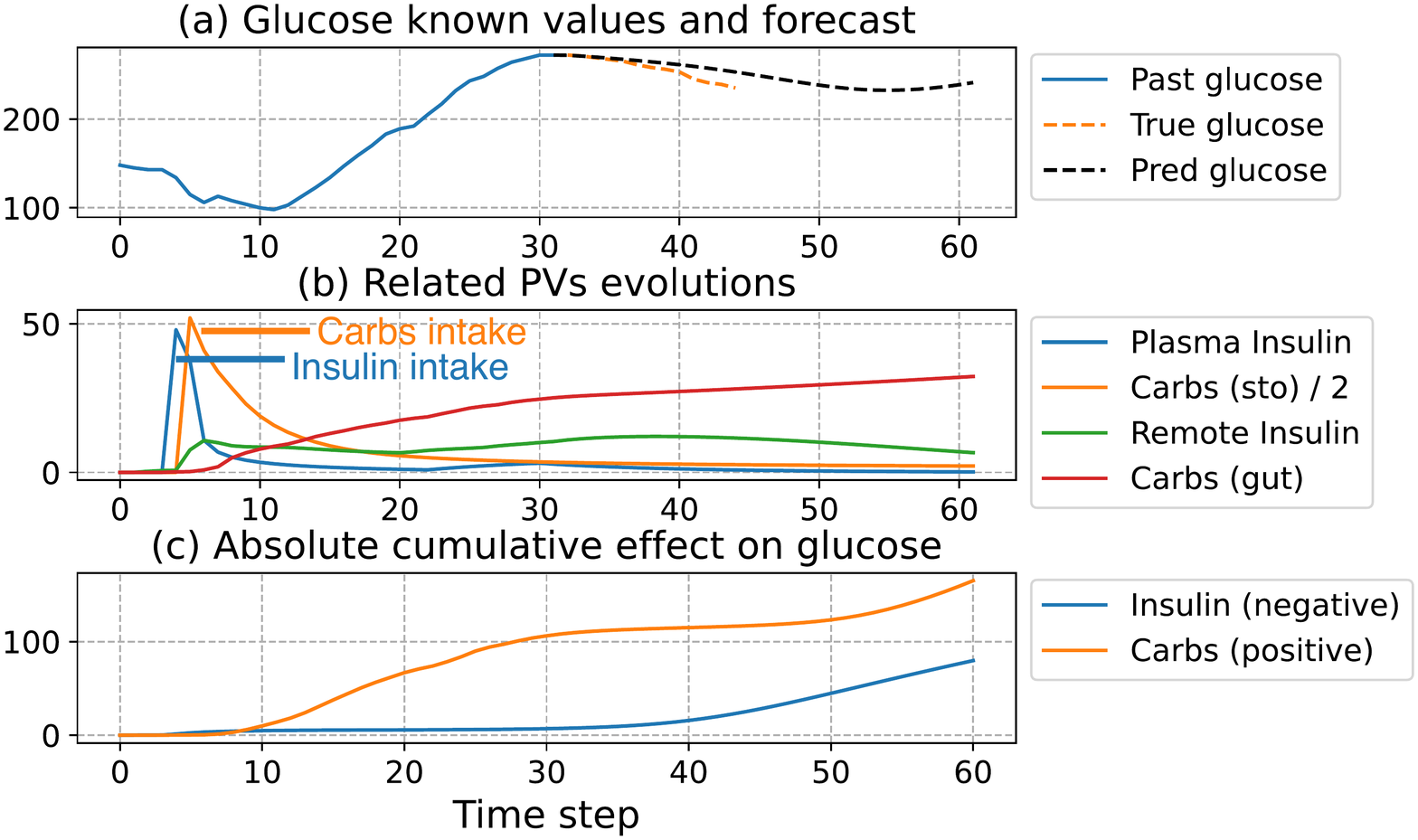}
\vspace{-10mm}
\caption{\label{fig:analysis}\textbf{HAD-Net parameters analysis.} Parallel analysis of glucose evolution (a), PVs evolution (b) and effects on glucose (c).}
\end{figure}

\paragraph{Glucose-level prediction by context.}
We analyze the performance of HAD-Net according to the physiological context of the patient. 
This helps to assess the clinical usefulness of our model in crucial contexts like after a meal intake (postprandial, post breakfast), 
or after insulin delivery (post bolus).
\figureref{fig:context_results} shows MAE (mean absolute error) at a prediction horizon of 30 minutes and the 50\% confidence interval.
As expected, we observe that glucose-level prediction overnight is an easier context compared to meal intake or insulin delivery. 
Moreover, HAD-Net still highlights competitive results in all contexts.
 \vspace{-5mm}
\paragraph{Analysis of HAD-Net parameters.}
Beyond predictions, we can visualize the evolution of physiological variables from HAD-Net. This helps to better interpret the insulin and carbs interactions, and their respective implications on the glucose level. \figureref{fig:analysis} depicts an example of physiological values calculated by HAD-Net, as well as the effect of insulin and carbs on the glucose level.
Here, the context is an insulin delivery (\figureref{fig:analysis}-b in blue) few minutes before a meal intake (\figureref{fig:analysis}-b in orange).
The glucose starts rising and only decreases after 30 time-steps. (\figureref{fig:analysis}-a). This is explained by the fact that the carbs have a short term impact on the glucose level while the insulin impact is delayed and slower (\figureref{fig:analysis}-c), where the absolute impact on glucose of a PV $p$ is $t \mapsto |\int_0^t I_{pG}(t)p(t)dt|$. These dynamics corroborate the physiological interactions that we set as the constraints of the model.
\vspace{-5mm}
\vspace{-10mm}
\section{Conclusion and Future Work}
\label{sec:conclusion}

HAD-Net simulates the diffusion of physiological quantities (glucose, insulin, carbs) into various body compartments (plasma, stomach, guts).
%
%
HAD-Net exhibits a good trade-off between accuracy and interpretability regarding glucose-level prediction. It achieves better forecast performances than shallow models and is as good as standard deep learning models. Beyond prediction, HAD-Net provides comprehensive diabetes-related insights that help to understand the metabolic specificity of each patient.
Future work calls for adapting the model to be more patient-specific, as well as integrating the kind of insulin (e.g. fast, ultra-fast). 
Also, HAD-Net is evaluated over patients with type-2 diabetes. It would be interesting to evaluate the relevance of our approach on type-1 that is more challenging.
\bibliography{ml4h_99_hadnet}

\appendix
\section{Details on GDPM}
\label{apd:gdpm}

\paragraph{Requirements} To rewrite a physiological model into a GDPM, we impose a few requirements. Most physiological models modeling diabetes fit with these requirements, or with minor adjustments as long as it stays consistent with reality. The reason for such a rewriting is that the diffusion magnitudes defined below are well suited to be learned by attention (see \sectionref{sec:model}) while keeping biological knowledge.

\paragraph{(1)} For all $t \geq 0$ and $i \in [1, K]$, we impose $p_i(t) \geq 0$. If a PV is always negative then we can simply consider its opposite instead. Yet, a variable whose sign is changing over time cannot fit with this model. Please note that this is very rare in diabetes physiological problems because we consider physical quantities such as insulin level or glucose level, which by definition are positive.

\paragraph{(2)} Every equation of the system should be decomposed into a sum of positive or negative functions depending on at most two variables. That is, $\forall 1 \leq i \leq K$,
\begin{equation}
    \frac{dp_i}{dt}(t) = \sum_{j=1}^K h_{ji}(p_j(t), p_i(t), t)
\end{equation} where $h_{ji}$ are functions that are either null, strictly positive or strictly negative and represent diffusion from $j$ to $i$. The sign constraints mean that a variable impact on another one has to be either constructive or destructive. We impose $(h_{ji} = 0 \text{ if } h_{ij} \neq 0)$, which means we model simple diffusion (not on both directions). The model can be generalized to support both directions diffusion if needed by having two matrices $\vec{F}$ instead of one. 

\paragraph{(3)} Diffusion magnitudes constraint: $\forall i, j,  I_{ij}(t) := |\frac{h_{ij}(p_i(t), p_j(t), t)}{p_i(t)} | \leq 1$ because we learn this function with attention scores that are less than 1 by definition. If the physiological model doesn't fit with this constraints, one can simply re-scale some variables. For most of the considered physiological models, $I_{ij}$ is a constant or have small variations, which makes it well-suited to be learned.

\paragraph{Definition}
We made the previous assumptions so that the model can be seen as the propagation of quantities along with a graph. For instance, the fact that insulin has a negative impact on glucose is modeled by a "negative" edge on a graph, meaning there is a diffusion from insulin to glucose that is destructive (insulin usage reduces glucose). Conversely, a "positive" edge means a variable consumption is transformed into another one, e.g. food transiting from the stomach to the gut. We call such a graph a graph diffusion physiological model (GDPM). Formally, a GDPM is defined by a set of $K$ nodes $p_1, p_2, ..., p_K$ -corresponding to PVs-, and weight edges:
\begin{equation}
    w_{ij}=
    \begin{cases}
      0, & \text{if}\ h_{ji} = 0 \\
      +1, & \text{if}\ h_{ji} > 0 \\
      -1, & \text{if}\ h_{ji} < 0 \\
    \end{cases}
  \end{equation}

The edges are thus oriented, and a non-negative weight indicates a diffusion direction. A weight is positive if variable $j$ has a constructive impact on the variable $i$, negative if it is destructive, and null if the variable $j$ doesn't affect directly the variable $i$. $w_{ij}$ represents the diffusion direction while the functions $I_{ij}$ defined above corresponds to diffusion magnitudes.

\section{Algorithms}

We provide pseudo-code for both the diffusion step (\algorithmref{alg:diffusion_step}) and the prediction (\algorithmref{alg:full}). Please note that we don't include $\Delta t$ in the algorithms below because it is a constant that can be considered included in $\vec{F}$.

\begin{algorithm2e}
 \KwIn{PVs $\vec{v}_t$, hidden state $\vec{H}_t$, model dimension $d$, default attention $a_0$, constraint matrix $\vec{A}$}
 \KwOut{increments $\vec{\Delta v}_{t}$, $\vec{\Delta H}_{t}$}
 $\vec{E}_t \gets \vec{v}_t^T || \vec{H}_t$\;
 $\vec{Q}, \vec{K}, \vec{V} \gets K\-Linear(\vec{E}_t)$ (3 times)\;
 $\vec{F} \gets \sigma ( a_0 + \frac{\vec{Q}\vec{K^T}}{d}  )$\;
 $\vec{M} \gets \vec{A} \odot \vec{F} - Diag((\vec{|A|}^T \odot \vec{F})_{\rightarrow})$\;
 $\vec{\Delta v_{t}} = \vec{M}\vec{v_{t}}$\;
 $\vec{\Delta H}_{t} = \vec{M}\vec{V}$\;
 \Return{$\vec{\Delta v}_{t}, \vec{H}_{\Delta t}$}
\caption{Diffusion step}
\label{alg:diffusion_step}
\end{algorithm2e}

\begin{algorithm2e}
 \KwIn{initial hidden state $\vec{H}_0$, model dimension $d$, default attention $a_0$, constraint matrix $\vec{A}$, measurements $\vec{X}_{1:w}$, window size $w$, prediction horizon $h$}
 \KwOut{glucose forecast $(V_{1, w+t})_{1 \leq t \leq h}$ where $V_{1, t}$ is the glucose PV}
 $\vec{v}_0 \gets (X_{1, 1}, 0, 0, ..., 0)$\;
 \For{$t \gets 1$ \textbf{to} $w$} {
  $\vec{v}_{t-1}^* \gets AddExogenous(\vec{v}_{t-1}, \vec{x}_{t})$\;
  $\vec{\Delta v}_{t-1}, \vec{\Delta H}_{t-1} \gets DiffusionStep(\vec{v}_{t-1}^*, \vec{H}_{t-1}, d, a_0)$\;
  $\vec{H}_t \gets BatchNorm(\vec{H}_{t-1} + \vec{\Delta H}_{t-1})$\;
  $\vec{v}_t \gets \vec{v}_{t-1}^* + \vec{\Delta v}_{t-1}$\;
  $\vec{v}_{t} \gets AddError(\vec{v}_{t}, \vec{x}_{t})$\;
  $\vec{v}_{t} \gets SetGlucose(\vec{v}_{t}, \vec{x}_{t})$\;
 }
 \For{$t \gets 1$ \textbf{to} $h$} {
  $\vec{\Delta v}_{w+t-1}, \vec{\Delta H}_{w+t-1} \gets DiffusionStep(\vec{v}_{w+t-1}, \vec{H}_{w+t-1}, d, a_0)$\;
   $\vec{v}_{w + t} \gets \vec{v}_{w + t-1} + \vec{\Delta v}_{w + t-1}$\;
  $\vec{H}_{w + t} \gets BatchNorm(\vec{H}_{w + t-1} + \vec{\Delta H}_{w + t-1})$\;
 }

 \Return{$(V_{1, w+t})_{1 \leq t \leq h}$}
 \caption{Prediction: Inference (first for loop) and Forecast (second for loop)}
 \label{alg:full}
\end{algorithm2e}

\section{Implementation details}
\label{apd:implementation}

\begin{figure}[hbt!]
\centering
\includegraphics[width=0.35\textwidth]{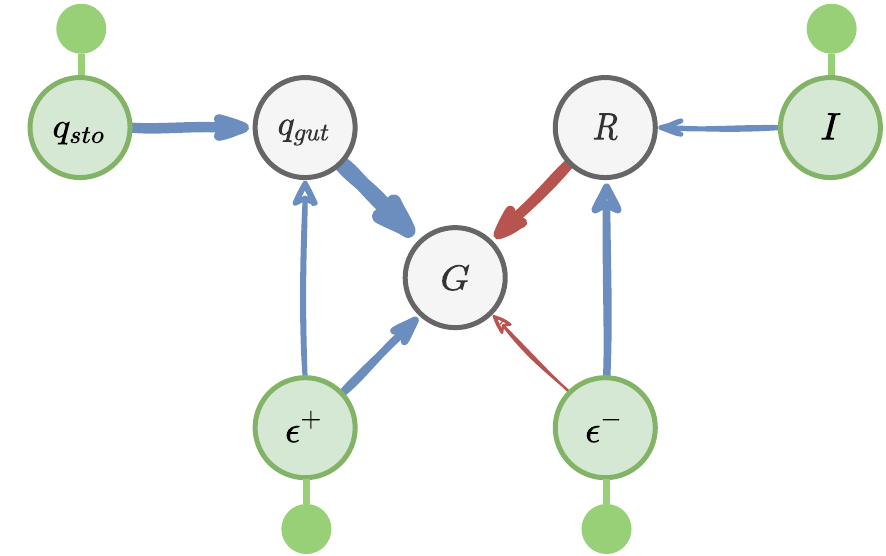}
\caption{\label{fig:model_diffusion}\textbf{Illustration of the diffusion process for $\vec{A}$ defined in \appendixref{apd:implementation}} Arrow colors correspond to the form of transformation, either constructive (blue arrows, $w_{ij} \geq 0$) or destructive (red arrows, $w_{ij} \leq 0$). Their size corresponds to magnitudes that vary over time and is made from attention. Green inputs mean a variable fills up the linked PV (i.e. during the operations AddExogenous or AddError). This is an example of a simple GDPM applied to our deep learning model. Other more complex GDPM can be considered.}
\end{figure}

The model parameters are $(\vec{H}_0, \vec{\Theta})$ where $\vec{\Theta}$ corresponds to the $3 \times K$ linear operator required to build the matrices Q, K, V (\sectionref{sec:diffusion_step}). Please note that $\vec{v}_0$ is not a parameter as we consider all PVs be to set at 0 at $t=0$, except glucose that is set to $X_{1, 1}$. We have chosen a model size $d = 32$ and we considered a physiological model such as $K = 7$. It gives a total number of $22k$ parameters. Depending on the dataset size, one can make a bigger model by using multiple layers instead of one in the attention process and also using many heads. Also, we have chosen a window size $w=32$ and a prediction horizon of $h=12$ steps with $\Delta t = 5$ minutes. It means we will provide continuous forecasts for up to one hour. We used the optimizer AdamW from PyTorch \citep{NEURIPS2019_9015} with a learning rate of $5.10^{-4}$ during 10 epochs. The dataset we considered is not very large, neither is our model, so the training took about a minute on CPUs. The matrix $\vec{A}$ we considered is the following.

\[
\vec{A} = \begin{pmatrix}
0 & 0 & -1 & 0 & 1 & 1 & -1 \\
0 & 0 & 0 & 0 & 0 & 0 & 0 \\
0 & 1 & 0 & 0 & 0 & 0 & 1 \\
0 & 0 & 0 & 0 & 0 & 0 & 0 \\
0 & 0 & 0 & 1 & 0 & 1 & 0 \\
0 & 0 & 0 & 0 & 0 & 0 & 0 \\
0 & 0 & 0 & 0 & 0 & 0 & 0
\end{pmatrix}
\]

where $\vec{v} = (G, I, R, q_{sto}, q_{gut}, \epsilon^+, \epsilon^-)$. G represents glucose, I plasma insulin, R remote insulin, $q_{sto}$ food in the stomach, $q_{gut}$ food in the gut, and $\epsilon^+, \epsilon^-$ the error compartments. Here $A_{1, 3} = -1$ means remote insulin decreases glucose level. This matrix was inspired and extracted from the physiological models of \citet{goel} and \citet{Bergman2005}. \figureref{fig:model_diffusion} illustrates this matrix by representing it as the corresponding GDPM with arbitrary weights.

\section{Compared models}
\label{apd:models}

\paragraph{Shallow models} We used the ARIMA model from the python package statsmodel \citep{seabold2010statsmodels} of order (5, 0, 0). Also, we have used the Ridge regressor from Scikit-learn \citep{scikit-learn} with the default parameters. The Gaussian Process was also built with Scikit-learn.

\paragraph{CNN-MLP} Our CNN is a stack of three PyTorch Conv1D of 3 channels and a respective kernel size of 7, 5, and 5. The three input channels correspond to the three times series of bolus intake, basal intake, and carbs intake. We flatten the resulting array and concatenate it with the glucose time series and its increments time series. We plug the latter in a MLP with hidden size 32 to project into a vector of the size of the desired horizons. MLP weights are trained on all patients and then retrained for each patient individually. Thus, this model is patient-specific.

\paragraph{Physiological model} The physiological model we used was implemented based on \citet{goel}.

\paragraph{GRU and LSTM} We add glucose increments to the time series of glucose, bolus, basal, and carbs intake. This input is fed into a GRU or LSTM of 2 layers, a hidden size of 64, and input size 5. The GRU or LSTM output is plugged in a 5 layers MLP that returns the forecasts for the provided horizons.

\end{document}